\documentclass[a4paper,conference]{IEEEtran}
\IEEEoverridecommandlockouts

\usepackage{cite}
\usepackage{amsmath,amssymb,amsfonts}
\usepackage{graphicx}
\usepackage{booktabs}
\usepackage{url}
\usepackage{algorithmic}
\usepackage{subcaption}

\renewcommand\IEEEkeywordsname{Keywords}

\begin{document}

\title{Contrail-to-Flight Attribution Using Ground Visible Cameras and Flight Surveillance Data}

\author{\IEEEauthorblockN{Ramon Dalmau, Gabriel Jarry \& Philippe Very}
\IEEEauthorblockA{Aviation Sustainability Unit (ASU) \\
EUROCONTROL\\
Brétigny-Sur-Orge, France}}

\maketitle

\begin{abstract}
Aviation's non-CO\textsubscript{2} effects, particularly contrails, are a significant contributor to its climate impact. Persistent contrails can evolve into cirrus-like clouds that trap outgoing infrared radiation, with radiative forcing potentially comparable to or exceeding that of aviation's CO\textsubscript{2} emissions. While physical models simulate contrail formation, evolution and dissipation, validating and calibrating these models requires linking observed contrails to the flights that generated them, a process known as contrail-to-flight attribution. Satellite-based attribution is challenging due to limited spatial and temporal resolution, as contrails often drift and deform before detection. In this paper, we evaluate an alternative approach using ground-based cameras, which capture contrails shortly after formation at high spatial and temporal resolution, when they remain thin, linear, and visually distinct. Leveraging the ground visible camera contrail sequences (GVCCS) dataset, we introduce a modular framework for attributing contrails observed using ground-based cameras to theoretical contrails derived from aircraft surveillance and meteorological data. The framework accommodates multiple geometric representations and distance metrics, incorporates temporal smoothing, and enables flexible probability-based assignment strategies. 
This work establishes a strong baseline and provides a modular framework for future research in linking contrails to their source flight.
\end{abstract}


\section{Introduction}\label{sec:intro}

Aviation contributes to climate change not only through CO\textsubscript{2} emissions but also through non-CO\textsubscript{2} effects, including nitrogen oxides (NO\textsubscript{x}), water vapour, aerosols, and contrails, ice-crystal clouds formed by aircraft at cruising altitudes. Persistent contrails can spread into extensive cirrus-like formations, trapping outgoing long-wave radiation and warming the planet. Recent studies suggest that the radiative forcing of contrail cirrus may be comparable to that of aviation CO\textsubscript{2} emissions~\cite{lee2021contribution,teoh2023global}, though the exact magnitude depends on the metric used~\cite{borella2024importance}.

Quantifying the climate impact of contrails, however, remains a major challenge. Their lifecycle depends on interconnected processes: ice nucleation, crystal growth, wind-driven dispersion, and interactions with natural clouds, all highly sensitive to atmospheric conditions. Small variations in temperature and relative humidity with respect to ice determine whether a contrail dissipates or persists. Combined with diurnal variability in radiative forcing (daytime cooling from solar reflection versus nighttime warming from infrared trapping), contrails’ net effect is strongly context-dependent and difficult to model reliably.

Most studies of contrail impacts rely on physical models such as CoCiP~\cite{schumann2012cocip} and APCEMM~\cite{fritz2020role}, which simulate contrail formation from aircraft parameters and weather conditions. These models are central to estimating contrails’ climate effects, but their accuracy is limited by uncertain inputs, especially relative humidity~\cite{gierens2020uncertainty}. Observations from satellites~\cite{chevallier2023linear} and ground-based imagery~\cite{van2025contrail} provide a crucial way to test and improve these models. Yet this requires one essential step: linking each observed contrail to the specific flight that produced it. Without such attribution, it is impossible to know whether a model’s predictions about contrail occurrence, persistence, or optical properties match reality. In short, contrail-to-flight attribution is what transforms raw imagery into actionable evidence for model calibration and validation.

Satellite imagery, however, has inherent limitations. By the time contrails are visible at satellite resolution, they have often drifted, deformed, or merged with other clouds, making alignment with flight trajectories highly uncertain. Ground-based cameras offer a promising alternative. They capture contrails just moments after formation, at high spatial and temporal resolution, when they are still thin, linear, and visually distinct. When paired with Automatic Dependent Surveillance–Broadcast (ADS-B) and meteorological data to simulate theoretical contrails, ground-based imagery opens a practical path toward precise contrail-to-flight attribution.

In this paper, we introduce a modular framework for contrail-to-flight attribution using ground-based visible cameras and evaluate it on the Ground Visible Camera Contrail Sequences (GVCCS) dataset~\cite{jarry_2025_15743988, jarry2025gvccs}. GVCCS contains video sequences recorded by a ground-based camera in Brétigny-sur-Orge, France, with detailed annotations for each contrail. Human labellers drew polygons directly on the images from the ground-based camera to capture contrail shapes, and each contrail was assigned a unique identifier that remains consistent across frames, enabling accurate tracking over time. Whenever possible, contrails were also linked to the flights that produced them, making the dataset ideally suited for testing and refining contrail-to-flight attribution algorithms.

The rest of this paper is structured as follows. Section~\ref{sec:soa} reviews related work on contrail-to-flight attribution. Section~\ref{sec:data} describes the data used in our study. Section~\ref{sec:matcher} presents the components of our modular framework, and Section~\ref{sec:method} details the experimental setup. Section~\ref{sec:results} reports the results, and Section~\ref{sec:conclusions} concludes with key findings and future directions.



\section{State of the Art}\label{sec:soa}

Physical models of contrails, such as CoCiP~\cite{schumann2012cocip} and APCEMM~\cite{fritz2020role}, provide detailed predictions of contrail formation and evolution, but their accuracy remains uncertain due to limited observational validation~\cite{gierens2020uncertainty}. Observational methods, in turn, detect contrails in satellite~\cite{chevallier2023linear} or ground-based~\cite{van2025contrail} imagery, but without a reliable way to attribute contrails to their source flights, these observations cannot be linked to flight-specific parameters and used for systematic model-observation comparison. Contrail-to-flight attribution is therefore a critical step for closing the gap between physics-based and observational approaches in contrail science.


Satellite imagery has long been the primary source of observational data, but it has inherent limitations when used for contrail-to-flight attribution. Geostationary satellites offer continuous temporal coverage (5–15~min refresh) but at relatively coarse spatial resolution (0.5–2~km), meaning contrails are only visible once they have grown thick and displaced under wind~\cite{itcovitz2024attribution}. Polar-orbiting satellites such as MODIS provide sharper images but only infrequent snapshots. In both cases, attribution becomes ambiguous as contrails drift, deform, or merge with surrounding clouds~\cite{geraedts2024contrails}. Furthermore, to the best of our knowledge, no human-labelled dataset exists that links contrails in satellite images to their source flights at scale, making direct evaluation impractical.

Synthetic datasets have been created to overcome this lack of ground truth. For instance,~\cite{chevallier2023linear} generated contrails using CoCiP and overlaid them on GOES-16 imagery, enabling the first systematic contrail instance segmentation experiments.~\cite{sarna2025benchmarking} extended this idea with SynthOpenContrails, which provides synthetic contrails linked to flights, creating the first benchmark for contrail-to-flight attribution algorithm comparison. While synthetic datasets are invaluable for algorithm development, their algorithmically generated ground truth prevents them from serving as a definitive benchmark for evaluation.

Ground-based cameras offer a complementary observational perspective. They capture contrails shortly after formation, when they remain thin, linear, and visually distinct. This high temporal and spatial resolution reduces attribution ambiguity and allows direct association with flight trajectory data (e.g., ADS-B)~\cite{low2025ground}. While their coverage is geographically limited, they provide the most practical basis for building human-labelled contrail-to-flight datasets, such as the GVCCS~\cite{jarry_2025_15743988, jarry2025gvccs}.

A range of algorithmic approaches have been explored for contrail-to-flight attribution. Geometric methods project ADS-B trajectories into image space and match them to detected contrails.~\cite{chevallier2023linear} combined Mask R-CNN detections with graph-based assignment strategies, incorporating wind-corrected (i.e., advected) trajectories.~\cite{geraedts2024contrails} employed scalable nearest-neighbour searches with penalties for temporal and spatial mismatch. These methods are interpretable but sensitive to trajectory uncertainties and contrail deformation.

Probabilistic approaches extend geometric matching by explicitly modelling uncertainty. For instance,~\cite{riggi2023ai} proposed Bayesian frameworks that account for uncertainties in both contrail geometry and flight tracks, improving robustness. Regional case studies such as~\cite{duda2024clusters} attributed MODIS-observed contrails to individual flights, quantifying cirrus coverage and radiative forcing, but relied on indirect physical consistency rather than labelled ground truth.

~\cite{sarna2025benchmarking} introduce CoAtSaC, a multi-frame, error-aware approach that attributes each detected (linearized) contrail to a flight by comparing the flight’s advected track to the contrail across multiple satellite frames.


\section{Data}\label{sec:data}

Our contrail-to-flight attribution framework relies on three key types of data: (1) contrails observed by a ground-based camera, (2) flight trajectory data for aircraft passing over the camera, and (3) meteorological measurements at cruising altitudes within the camera’s field of view, used to simulate theoretical contrails. In this section, we describe each of these essential data modalities and explain how they are implemented in our study.

\subsection{Observed Contrails}

The primary element of our contrail-to-flight attribution framework is the set of visible contrail instances captured in ground-based camera imagery. To develop and validate our attribution algorithm, we leverage a labelled benchmark that provides human-verified contrail-to-flight pairings. Specifically, we use the GVCCS dataset~\cite{jarry_2025_15743988, jarry2025gvccs}, a curated collection of approximately 24K annotated frames from 122 videos.

Each frame includes multi-polygons delineating visible contrails in pixel coordinates. These annotations capture both newly formed contrails, appearing shortly after formation directly above the camera, and older contrails that originated outside the field of view but drifted into it due to wind advection. Each contrail is manually labelled with a unique identifier and a status tag (“new” or “old”), ensuring temporal consistency across frames and enabling tracking of individual contrails as they evolve. A “new” contrail refers to a contrail that is first observed forming within the camera’s field of view and that can be reliably associated with the flight that generated it. In contrast, an “old” contrail corresponds either to a contrail advected into the camera’s field of view from outside, or to a potentially new contrail that could not be linked to a specific flight (e.g. due to cloud occlusion or other visual limitations). This distinction also enables quantitative evaluation of contrail-to-flight attribution algorithms.

\begin{figure*}[!ht]
    \centering
    \begin{subfigure}{0.25\textwidth}
      \includegraphics[width=\linewidth]{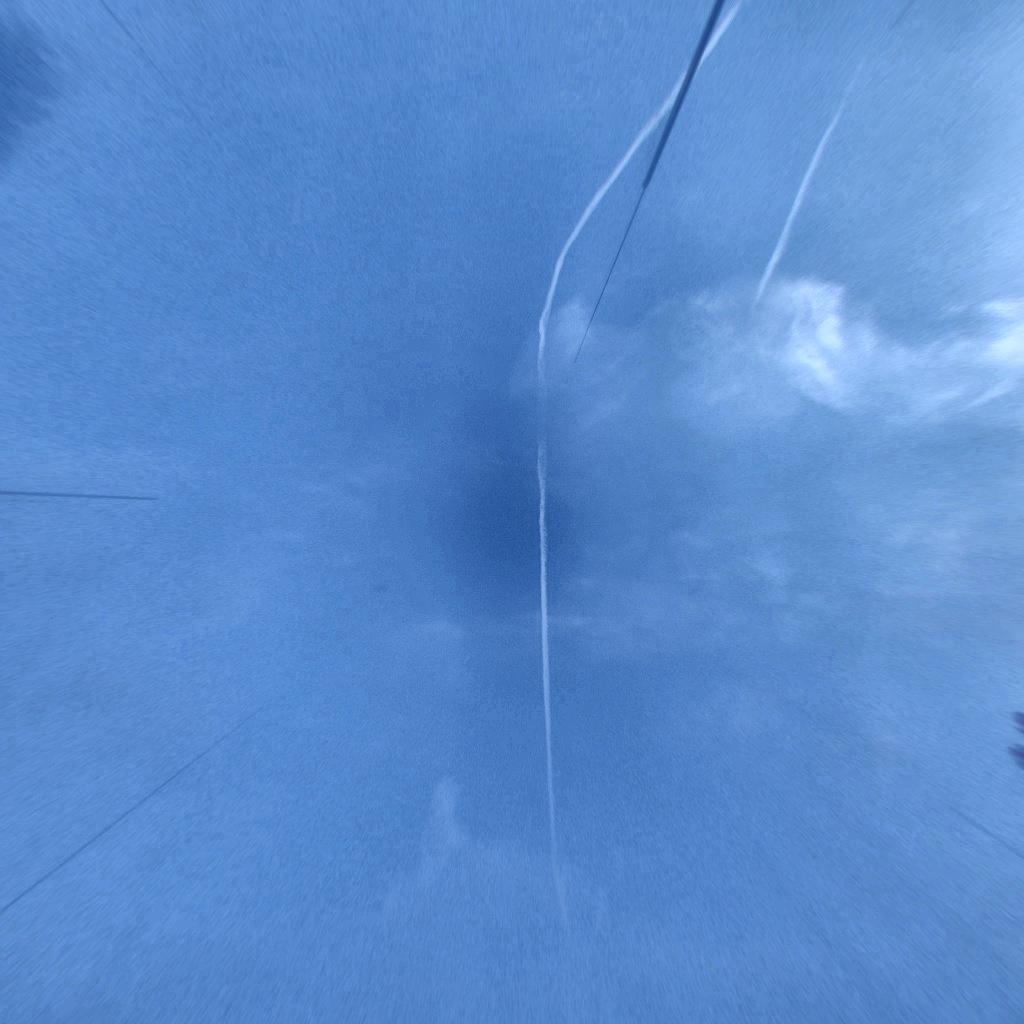}
      \caption{04:45:30}
      \label{fig:raw_044530}
    \end{subfigure}
    \begin{subfigure}{0.25\textwidth}
      \includegraphics[width=\linewidth]{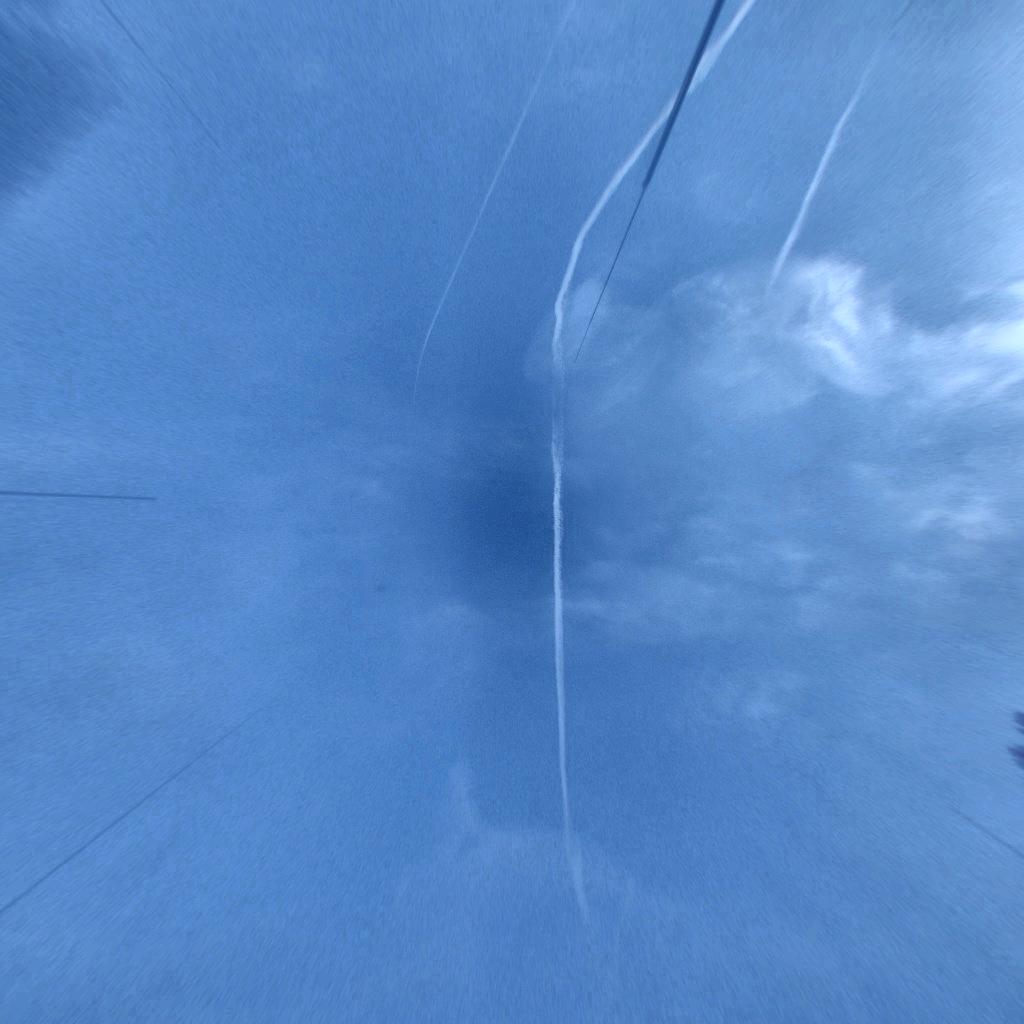}
      \caption{04:48:30}
      \label{fig:raw_044830}
    \end{subfigure}
    \begin{subfigure}{0.25\textwidth}
      \includegraphics[width=\linewidth]{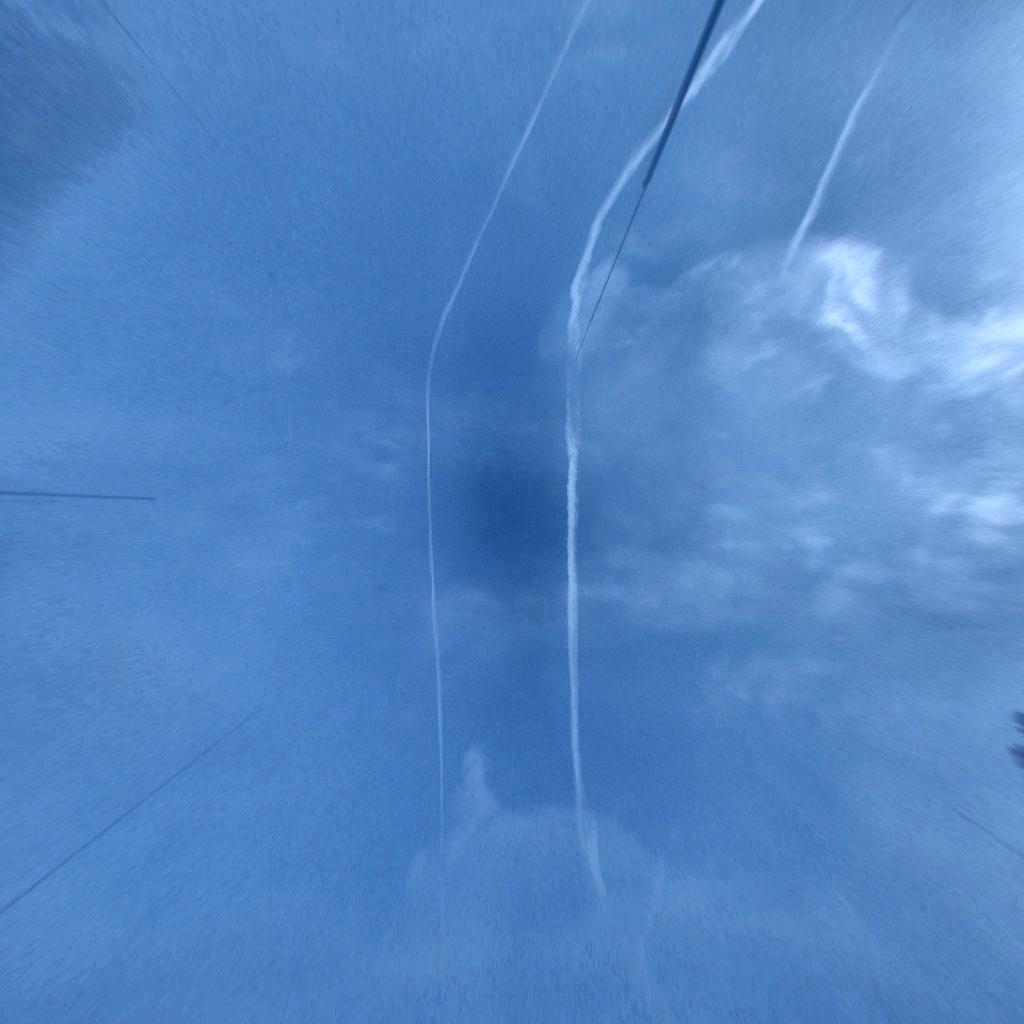}
      \caption{04:51:30}
      \label{fig:raw_045130}
    \end{subfigure}
    \caption{Illustrative sequence of frames on 26 April 2024.}
    \label{fig:raw}
\end{figure*}

\begin{figure*}[htbp!]
    \centering
    \begin{subfigure}{0.25\textwidth}
      \includegraphics[width=\linewidth]{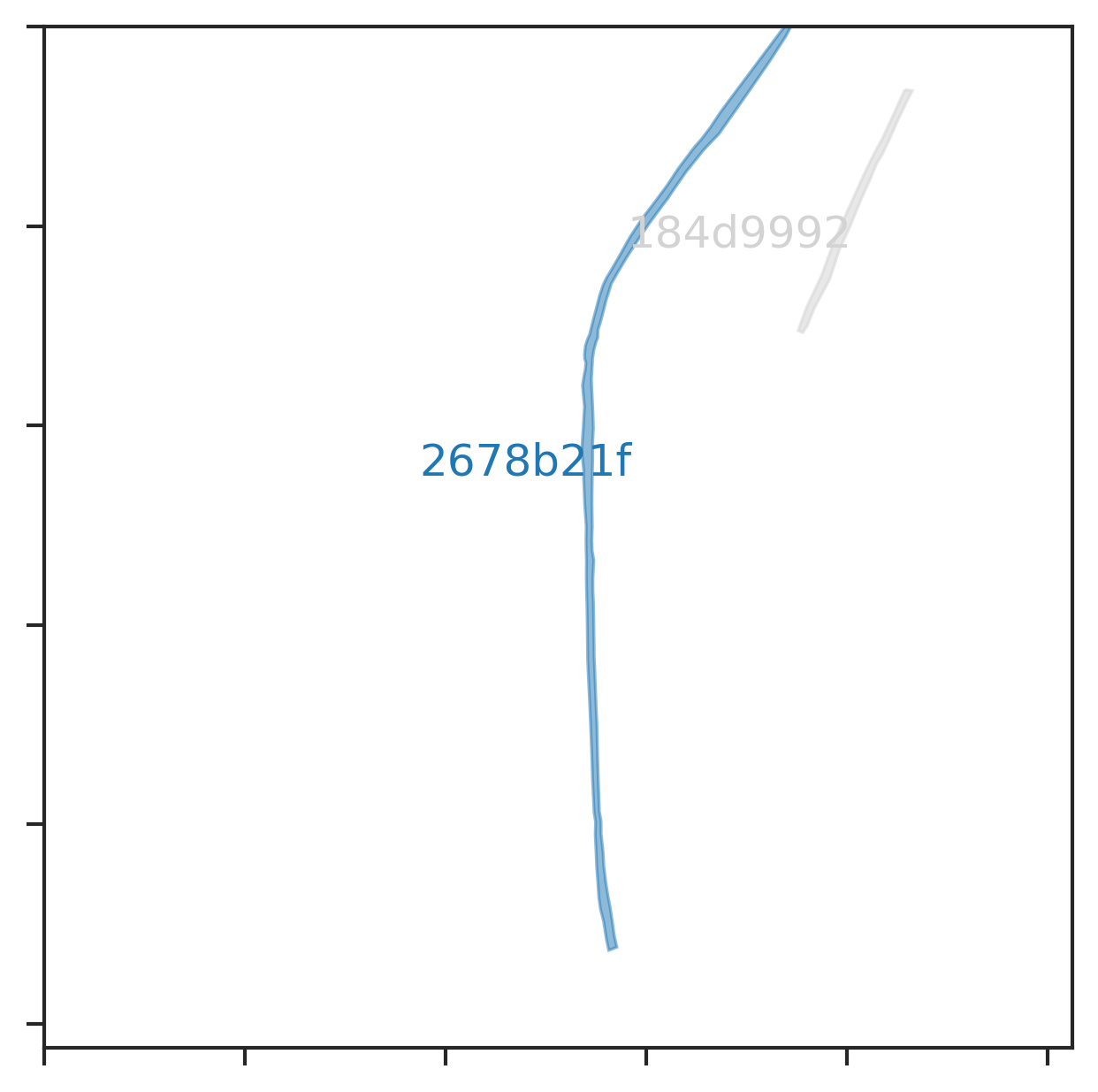}
      \caption{04:45:30}
      \label{fig:ann_044530}
    \end{subfigure}
    \begin{subfigure}{0.25\textwidth}
      \includegraphics[width=\linewidth]{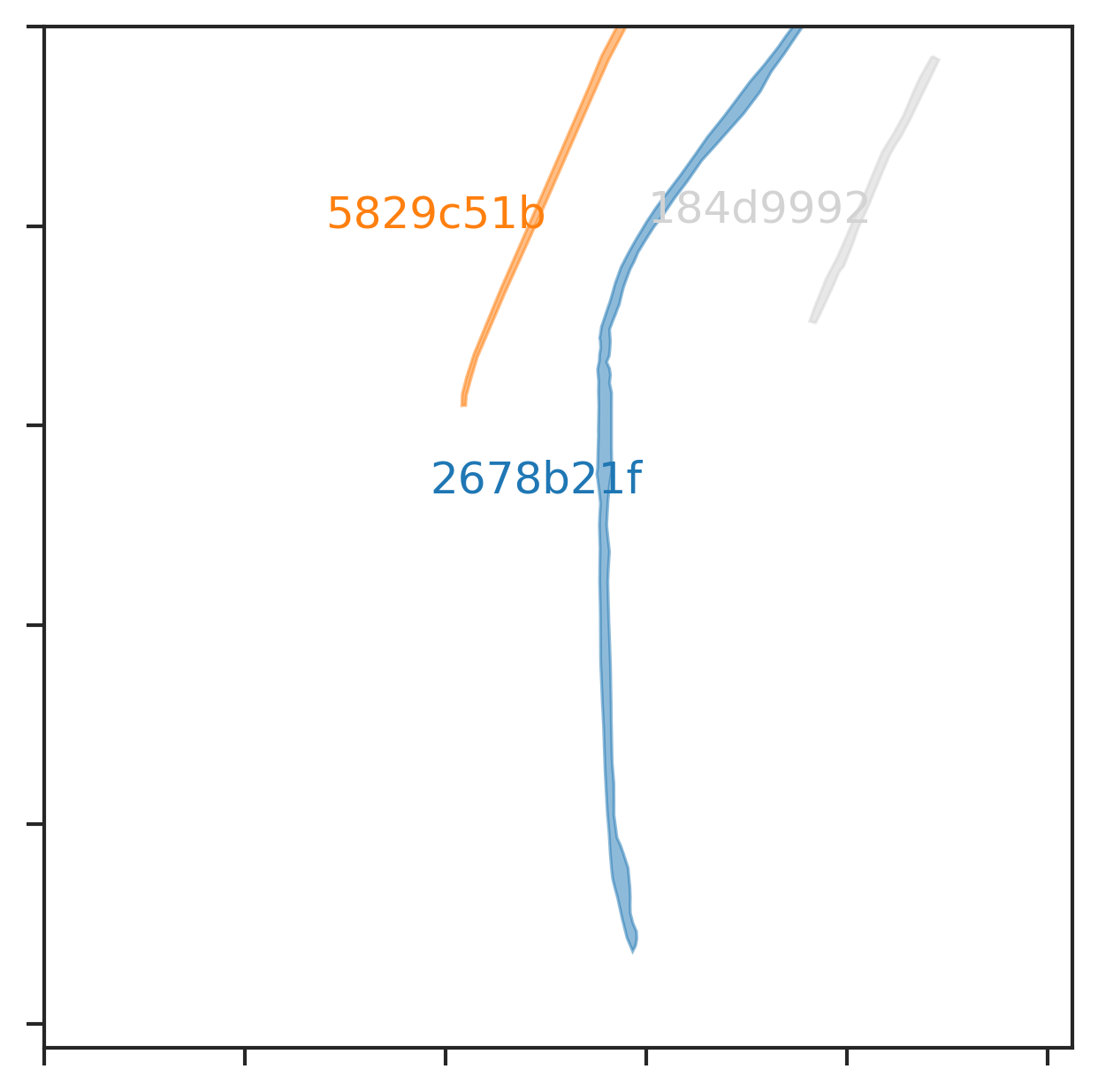}
      \caption{04:48:30}
      \label{fig:ann_044830}
    \end{subfigure}
    \begin{subfigure}{0.25\textwidth}
      \includegraphics[width=\linewidth]{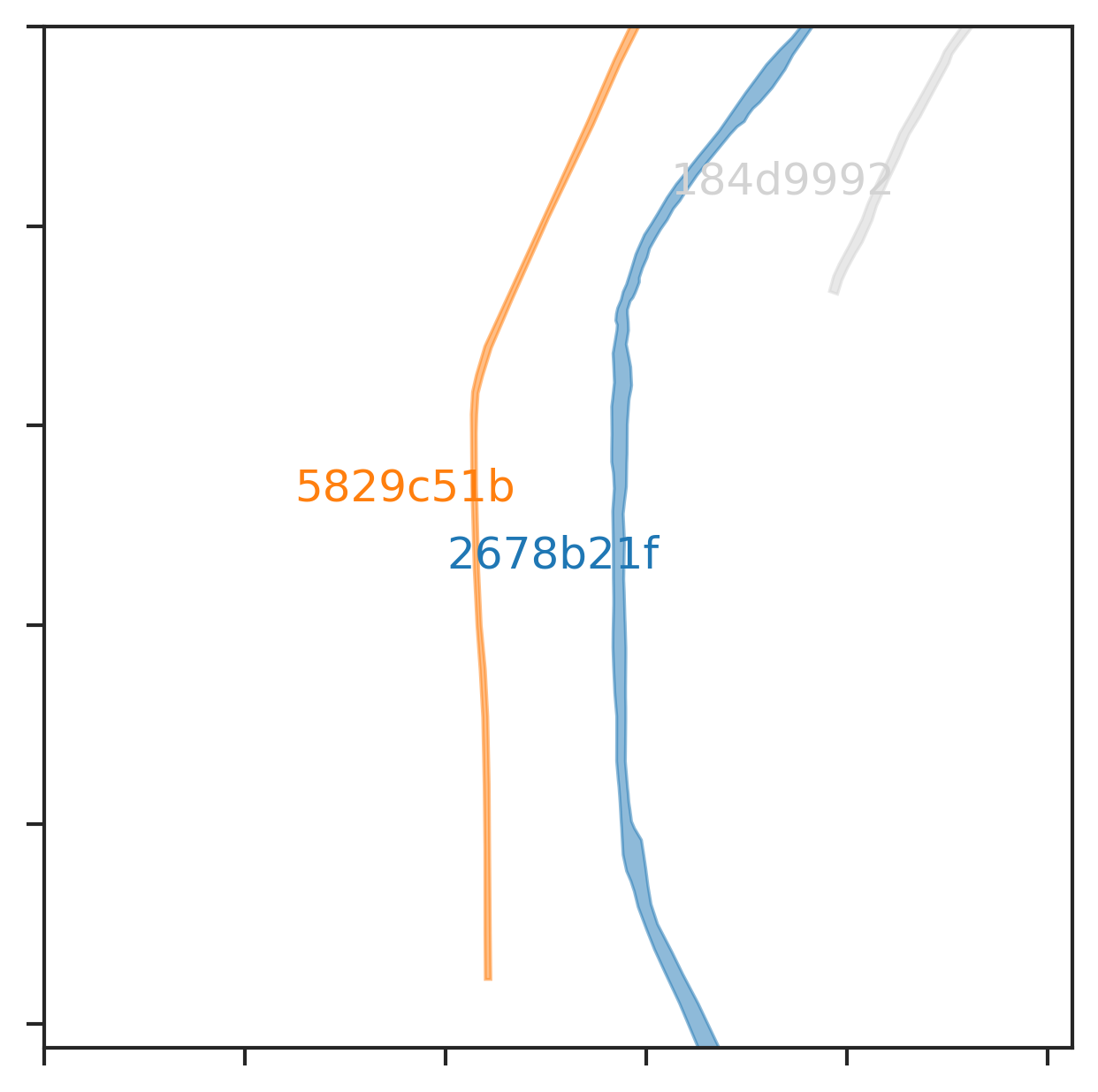}
      \caption{04:51:30}
      \label{fig:ann_045130}
    \end{subfigure}
    \caption{Annotations for the illustrative sequence shown in Fig.~\ref{fig:raw}. The sequence includes two new and one old contrails.}
    \label{fig:annotations}
\end{figure*}

Figure~\ref{fig:raw} presents a sample image sequence from GVCCS, recorded on 26~April~2024. The GVCCS operates at a  temporal resolution of 30~s. For improved clarity and to better illustrate the contrail evolution, however, Figure~\ref{fig:raw} displays one frame every 3~minutes. Figure~\ref{fig:annotations} illustrates the human-generated annotations for the same sequence. New contrails are colour-coded and retain consistent identifiers across frames, while old contrails are shown in grey.

\subsection{Flight Trajectory Data}

The GVCCS dataset also includes the flight trajectories (ADS-B observations) for all aircraft passing above the camera during each observation sequence. Crucially, these trajectories are linked to the contrail annotations through consistent flight identifiers, enabling direct and rigorous comparison between observed contrails and their originating flights.

For each frame, candidate flight positions are projected into the pixel coordinate system, allowing geometric comparisons with observed contrails. The dataset also includes metadata such as callsigns and aircraft types, which can support filtering (e.g., excluding propeller aircraft) or computing theoretical contrail predictions with physics-based models like CoCiP.

\subsection{Meteorological Data}

Reliable contrail-to-flight attribution depends critically on the atmospheric conditions that control a contrail's formation, persistence, and transport. To model this, our method leverages ERA5 reanalysis data from the European Centre for Medium-Range Weather Forecasts (ECMWF). We extract key meteorological fields, including horizontal winds, temperature, pressure, and relative humidity, at isobaric levels between 200 and 300 hPa, encompassing the standard cruising altitudes.

These fields are the essential inputs for simulating contrail advection and evolution. By driving physical models (e.g., CoCiP) or simpler Lagrangian advection models, the ERA5 data allows us to generate a set of plausible contrail trajectories. 

For each observed contrail, the framework compares it with simulated contrails generated from candidate flight trajectories in order to identify its most probable source flight. Although we rely on ERA5 for its high spatiotemporal resolution and public availability, the framework is compatible with any gridded meteorological dataset of sufficient quality.

\begin{figure*}[htbp!]
    \centering
    \begin{subfigure}{0.25\textwidth}
      \includegraphics[width=\linewidth]{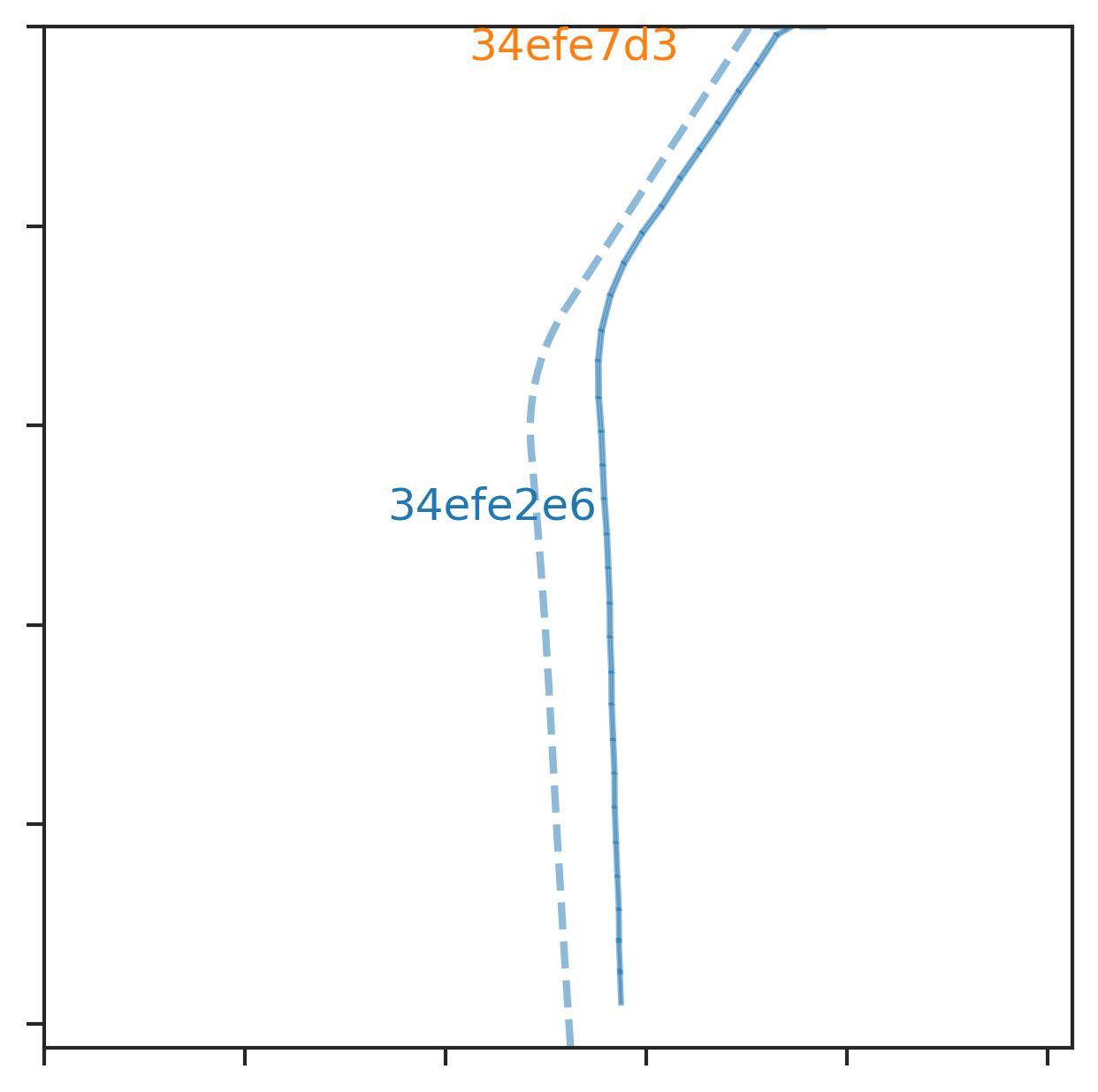}
      \caption{04:45:30}
      \label{fig:dry_044530}
    \end{subfigure}
    \begin{subfigure}{0.25\textwidth}
      \includegraphics[width=\linewidth]{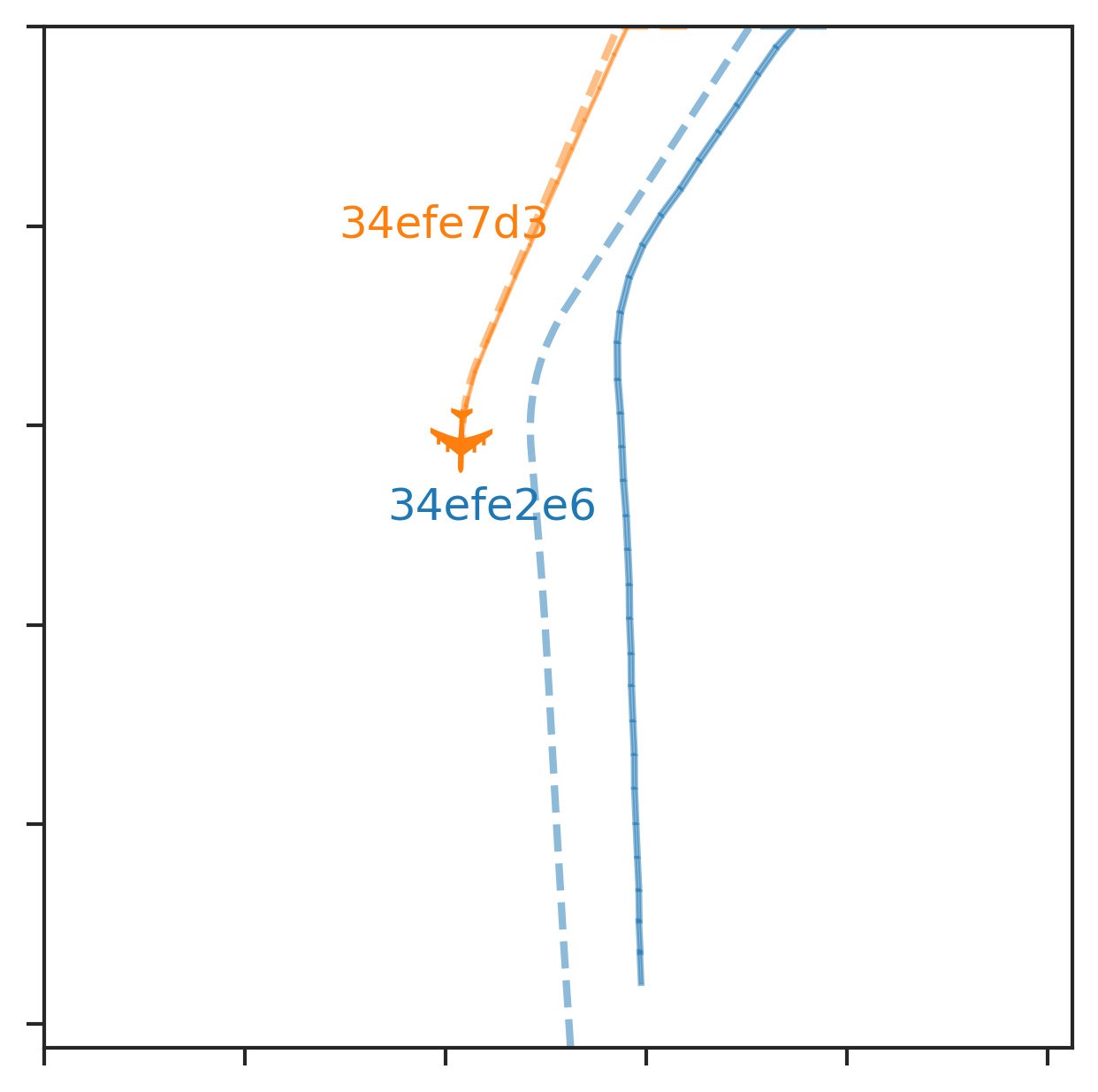}
      \caption{04:48:30}
      \label{fig:dry_044830}
    \end{subfigure}
    \begin{subfigure}{0.25\textwidth}
      \includegraphics[width=\linewidth]{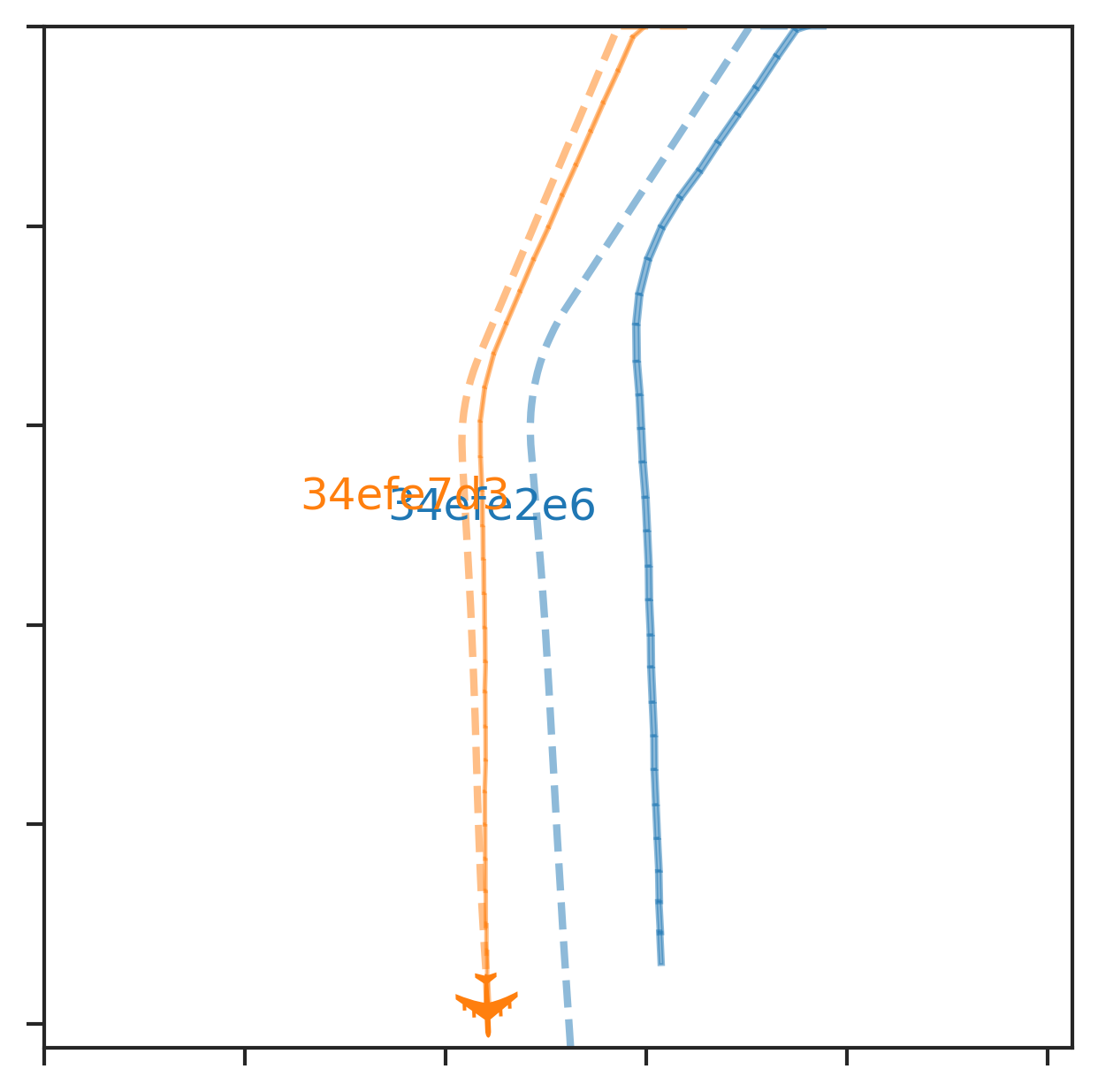}
      \caption{04:51:30}
      \label{fig:dry_045130}
    \end{subfigure}
    \caption{Theoretical contrails generated by a simple dry advection model for the flights crossing the camera’s field of view during the sequence shown in Fig.~\ref{fig:raw}. Dashed lines indicate the actual flight trajectories, while solid multi-polygons represent the simulated contrails.}
    \label{fig:advected}
\end{figure*}

Figure~\ref{fig:advected} illustrates the theoretical contrails generated by a simple dry advection model for the flights  crossing the camera’s field of view during the sequence shown in Fig.~\ref{fig:raw}. Dashed lines indicate the actual flight trajectories, while the solid multi-polygons represent the corresponding simulated contrails. Note that when the width of a contrail is very small, the multi-polygons may appear as simple lines.

\section{Contrail-to-Flight Attribution Framework}\label{sec:matcher}

An aircraft emits exhaust continuously along its trajectory. The exhaust released at a given point is transported and deformed by the wind, so an observed contrail at time $t$ generally corresponds to exhaust emitted earlier, at time $t' < t$, and advected to a different location. We formulate contrail-to-flight attribution as the problem of matching observed contrails with theoretical advected contrails generated by flights passing above the camera. Our framework addresses this task by: (i) restricting comparisons to spatio-temporally plausible pairs, (ii) computing geometric distances, (iii) aggregating distances over time for robustness, (iv) mapping distances to probabilities, and (v) solving the resulting assignment problem.


\subsection{Temporal Filtering}\label{sec:temporal_filtering}

Let $c_i$ denote an observed contrail with formation time $t_i$, defined as the first time when this contrail identifier was observed. Consider a candidate flight $f_j$ that generates an advected plume (i.e., a theoretical contrail) composed of geometries $\{g_{j1}, \dots, g_{jk}\}$ with corresponding formation times $\{t_{j1}, \dots, t_{jk}\}$. Using CoCiP or a dry advection model\footnote{https://py.contrails.org/notebooks/advection.html}, for example, each geometry is represented as a simple rectangle, determined by the segment length, width, and orientation.  

To retain only the relevant segments of the entire contrail, we introduce two temporal tolerance parameters, $\Delta t_{\text{before}}$ and $\Delta t_{\text{after}}$, which define the acceptable time window around $t_i$. A segment $g_{jk}$ is retained if its formation time satisfies:

\begin{equation}
    t_i - \Delta t_{\text{before}} \le t_{jk} \le t_i + \Delta t_{\text{after}}.
\end{equation}

If no segments satisfy this condition, the pair $(c_i, f_j)$ is discarded, as the flight is unlikely to have produced the observed contrail. If multiple segments survive, their unary union is computed to form a single composite geometry. This ensures that the full spatial extent of the plume relevant to $c_i$ is considered when computing geometry-based distance metrics, such as overlap or Hausdorff distance, for attribution purposes.

\subsection{Geometric Distance}\label{subsec:geometric_distance}

For each temporally compatible pair $(c_i, f_j)$, a non-negative geometric distance $d_{ij}$ is computed, where smaller values indicate better agreement between the observed and theoretical contrails. The choice of distance metric depends on the geometry representation of the contrails. For example, if both the observed and theoretical contrails are represented as polygons, the Intersection over Union (IoU) can be used. All distances are computed in the pixel coordinate system. To prevent assigning probability mass to implausible matches, distances exceeding a physically meaningful cut-off are excluded. Specifically, we introduce a threshold $\tau_d$ representing the maximum acceptable separation after accounting for advection uncertainty. If $d_{ij} > \tau_d$, we set $d_{ij} = \infty$.

\subsection{Memory-Aware Aggregation}
\label{subsec:memory_aggregation}

Instantaneous pairwise distances may fluctuate due to intermittent detections or advection mismatches. To stabilize the attribution decisions, a memory mechanism aggregates evidence over time. Specifically, this mechanism maintains a running summary of past distances between each observed and theoretical contrail pair. For example, for each pair $(c_i,f_j)$, one can compute an exponentially weighted moving average (EWMA) of the instantaneous distance $d^{(t)}_{ij}$ observed at $t$:

\begin{equation}
    \tilde D_{ij}^{(t)} \;=\; \alpha\, \tilde D_{ij}^{(t-1)} \;+\; (1-\alpha)\, d^{(t)}_{ij}, 
    \qquad \alpha \in [0,1].
\end{equation}

Larger values of $\alpha$ emphasize persistence, resulting in stronger smoothing over time, whereas smaller values allow the memory to adapt more quickly to changes in distance. 

\subsection{Distances to Probabilities}
\label{subsec:distance_to_probabilities}

The next step is to convert geometric distances into probabilities. Given the distance matrix $\tilde D \in \mathbb{R}^{m \times n}$ of $m$ contrails and $n$ flights, we first define scores $S_{ij} = -\tilde D_{ij}$ for finite entries, assigning a very negative sentinel value to excluded pairs. Each row is then converted into a probability distribution using a softmax with inverse temperature 
$\beta > 0$:

\begin{equation}
    P_{ij} \;=\; \frac{\exp\big(\beta S_{ij}\big)}{\sum_{k=1}^n \exp\big(\beta S_{ik}\big)}.
\end{equation}

To suppress low-confidence associations, a probability floor $\tau_p$ is applied: if $P_{ij} < \tau_p$, we set $P_{ij} = 0$ for assignment purposes. The inverse temperature $\beta$ controls the selectivity of the mapping, with larger values producing more peaked distributions. Note that the probability distribution can be computed either row-wise or over the entire matrix, depending on the choice of denominator aggregation. This choice primarily depends on the subsequent assignment algorithm.

\subsection{Assignment}\label{subsec:assignment}

Given the probability matrix $P \in [0,1]^{m \times n}$, the objective is to select the most likely source flight for each observed contrail. This can be done in several ways: independently for each row, or as a one-to-one assignment over the entire matrix using the Hungarian algorithm, for instance. 




For each contrail $c_i$, the framework returns the selected flight identifier (if any), along with the corresponding attribution probability $P_{ij}$ and the aggregated distance $D_{ij}^{(t)}$.

\section{Particular Implementation}\label{sec:method}

In this section we instantiate the general attribution framework introduced above. While the framework is designed to be modular, here we describe the concrete design  choices we implemented for evaluation. Each step of the pipeline is specified: how observed and theoretical contrails are represented, how spatial distances are  computed and temporally aggregated, and finally how assignments are made. 

\subsection{Geometries}

The first step is to represent both observed and theoretical contrails in a form suitable for spatial comparison.

\subsubsection{Annotations}

Observed contrails are provided as human- or machine-generated annotations in the form of polygon vertices. In our implementation, each polygon is reduced to its central path by applying a morphological thinning operation, which produces a one-pixel-wide structure along the centerline of the shape. The resulting skeleton is represented as a graph of connected pixels, from which we extract the longest path to serve as the contrail geometry.
 
\begin{figure}[htbp!]
    \centering
    \begin{subfigure}{0.4\textwidth}
      \includegraphics[height=\linewidth, angle=90]{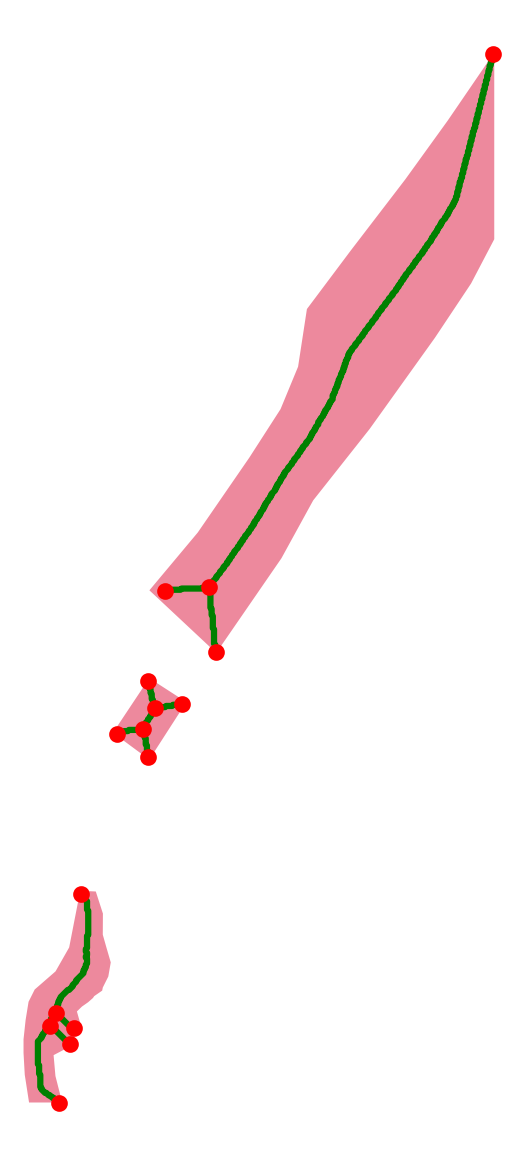}
      \caption{Graph of the geometry skeleton.}
      \label{fig:geometry_graph}
    \end{subfigure}
    \begin{subfigure}{0.4\textwidth}
      \includegraphics[height=\linewidth, angle=90]{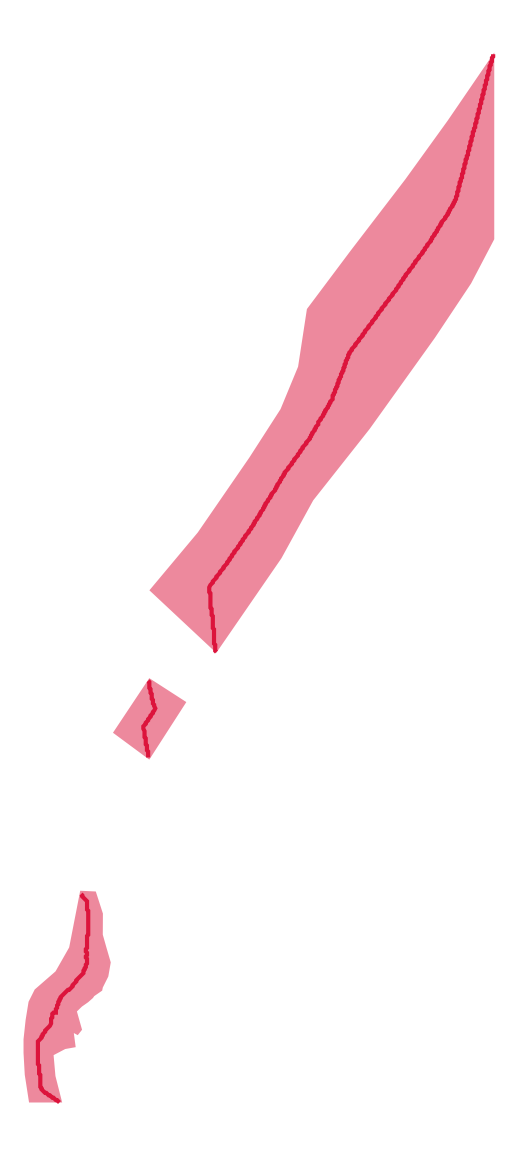}
      \caption{Longest path in the graph.}
      \label{fig:geometry_skeleton}
    \end{subfigure}
    \caption{Conversion of an annotated contrail into a geometry.}
    \label{fig:geometry}
\end{figure}

Figure~\ref{fig:geometry} illustrates the process: (1) the skeleton of the contrail mask is converted into a graph, and (2) the longest path is extracted to represent the contrail geometry. Each polygon is processed independently in the case of multi-polygon annotations.

\subsubsection{Theoretical Contrails}

Theoretical contrails are computed using dry advection with an initial width of 100\,m. Every 10 seconds, we assume the aircraft emits new exhaust. Each exhaust parcel is then propagated forward using the \texttt{pycontrails} dry advection model with a 30-second integration step. This model simulates how the plume moves with the wind (horizontally and vertically) and broadens with time.  

At each integration step and for each waypoint, the model outputs a segment characterized by its position, length, orientation, and width, which together define a polygon. Each polygon is also timestamped by its formation time. For temporal filtering, observed contrails are therefore compared only to the union of polygons with compatible formation times.  

\subsection{Distance Metric}

To compare an observed contrail with a theoretical contrail, we use a directed (asymmetric) variant of the Hausdorff distance. Let \(A\) denote the observed contrail geometry (i.e., its centerline) and \(B\) the union of polygons from the theoretical contrail. The distance from \(A\) to \(B\) is defined as:

\begin{equation}
d(A \to B) \;=\; \max_{a \in A} \min_{b \in B}\lVert a - b\rVert.
\end{equation}

That is, for each point \(a \in A\) we measure the Euclidean distance to the nearest point in \(B\), and we then take the maximum of these nearest distances.  

Intuitively, this measures how well the observed contrail is covered by the predicted contrail: if every point of \(A\) lies close to some point of \(B\), the distance is small; if even one part of the observation is far from the plume, the distance is large. 


Distances exceeding $\tau_d = 30$ pixels are discarded as implausible and replaced with $\infty$, preventing spurious matches.  

To stabilize the distance signal and improve assignment robustness, we employ an EWMA with $\alpha = 0.7$, balancing persistence with responsiveness to evolving contrail shapes.

\subsection{Assignment Strategy}

Finally, we assign each observed contrail to the most likely theoretical source. In this implementation, we use a simple greedy strategy: for each observed  contrail, we select the candidate flight with the highest non-zero probability using $\beta = 1$.  Each row is processed independently, without enforcing one-to-one constraints. Although this does not guarantee global optimality, it is computationally efficient and works well in practice when contrails have clearly dominant matches.  

Because assignment is done per row, we normalize probabilities row-wise, so the sum across candidates for each observed contrail equals one. Furthermore, to increase robustness, we apply a threshold of $\tau_p = 0.5$: only assignments with   probability above 50\% are accepted. Observed contrails with no confident match are left unassigned to avoid false attributions.

\section{Results}\label{sec:results}

For our experiments, we used the full GVCCS dataset. 
To isolate the performance of the attribution algorithm from the quality of contrail detection, we treated the provided polygon annotations as perfect detections. This allowed us to focus exclusively on evaluating contrail-to-flight attribution performance. Each newly forming contrail in the dataset is annotated with the flight identifier of the aircraft that generated it. We used this information as ground truth for attribution.

Annotations are categorized as either \emph{new} or \emph{old} contrails. After processing with the contrail-to-flight attribution algorithm, each contrail falls into one of two main categories:

\begin{itemize}
 \item \textbf{Attributed contrail:} the algorithm assigned a flight to the contrail.
 \item \textbf{Unattributed contrail:} no flight was assigned.
\end{itemize}

For attributed contrails, the outcomes are classified as:

\begin{itemize}
 \item \textbf{Correct Attribution:} the contrail was attributed to the correct flight.
 \item \textbf{Wrong Attribution:} the contrail was attributed to an incorrect flight.
 \item \textbf{False Attribution:} the contrail was actually an old contrail.
\end{itemize}

For unattributed contrails, the possible outcomes are:

\begin{itemize}
 \item \textbf{Correct Omission:} the contrail was an old contrail.
 \item \textbf{Missed Attribution:} the contrail was new but was not attributed.
\end{itemize}

This section presents the outcomes of our experimental evaluation. We first provide individual examples to illustrate the possible results of the attribution algorithm, followed by an analysis of its aggregated performance on the full dataset.

\subsection{Illustrative Examples}

In the figures below, following the conventions established in previous figures, old contrails are shown in grey. The actual flight paths are indicated with dashed lines, while the theoretical contrails are represented as solid polygons, which may appear as lines when the advected contrail is thin. Annotations are displayed as polygons, even if internally represented as lines by the algorithm, and are coloured according to the attributed flight, if any.  These examples highlight the different ways the algorithm can attribute or omit contrails, providing visual insight into its decision-making process.

\begin{figure}[htbp!]
    \centering
    \begin{subfigure}{0.5\textwidth}
      \includegraphics[width=\linewidth]{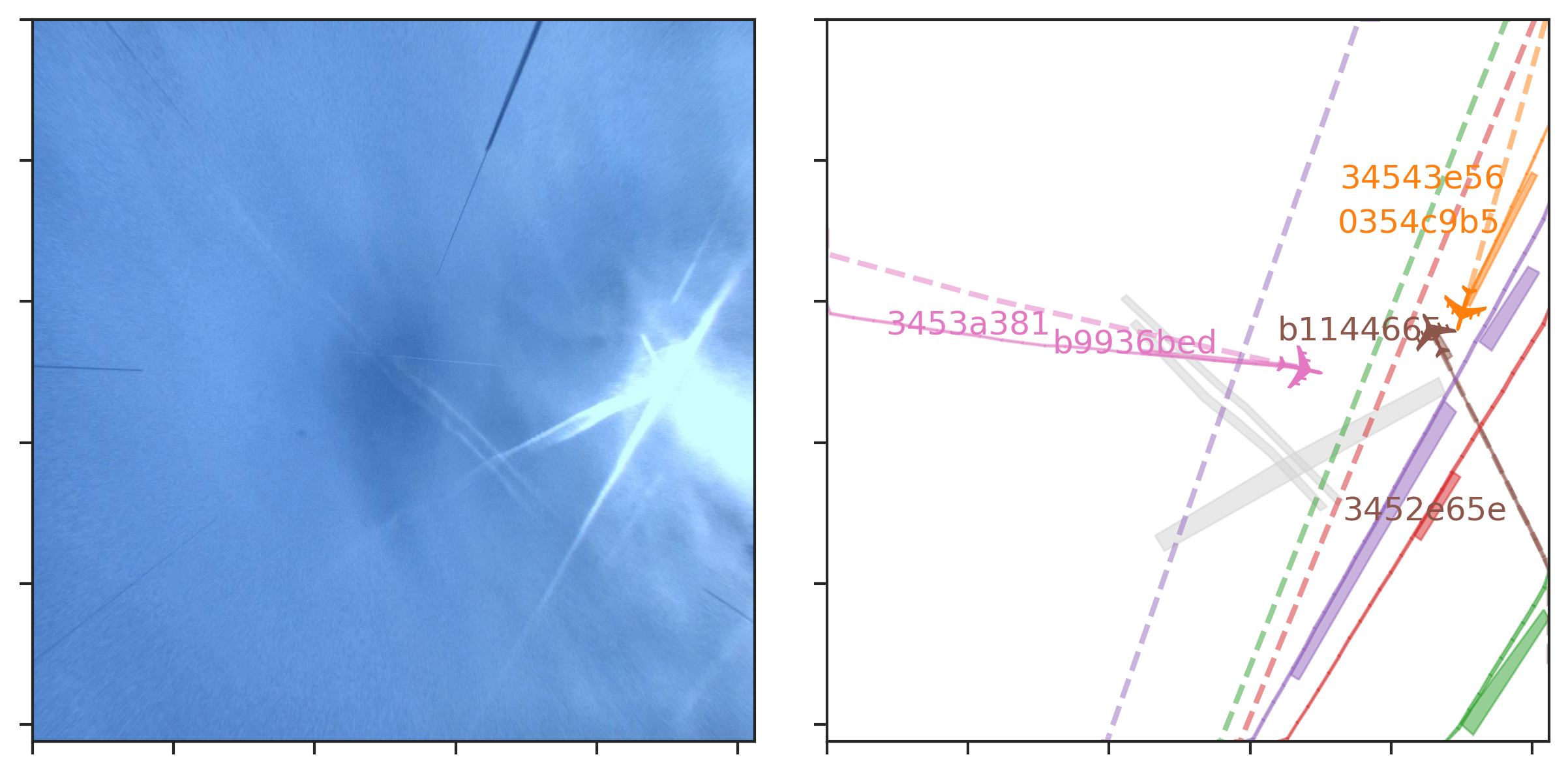}
      \caption{Correct Attribution}
      \label{fig:attr_correct}
    \end{subfigure}
    \begin{subfigure}{0.5\textwidth}
      \includegraphics[width=\linewidth]{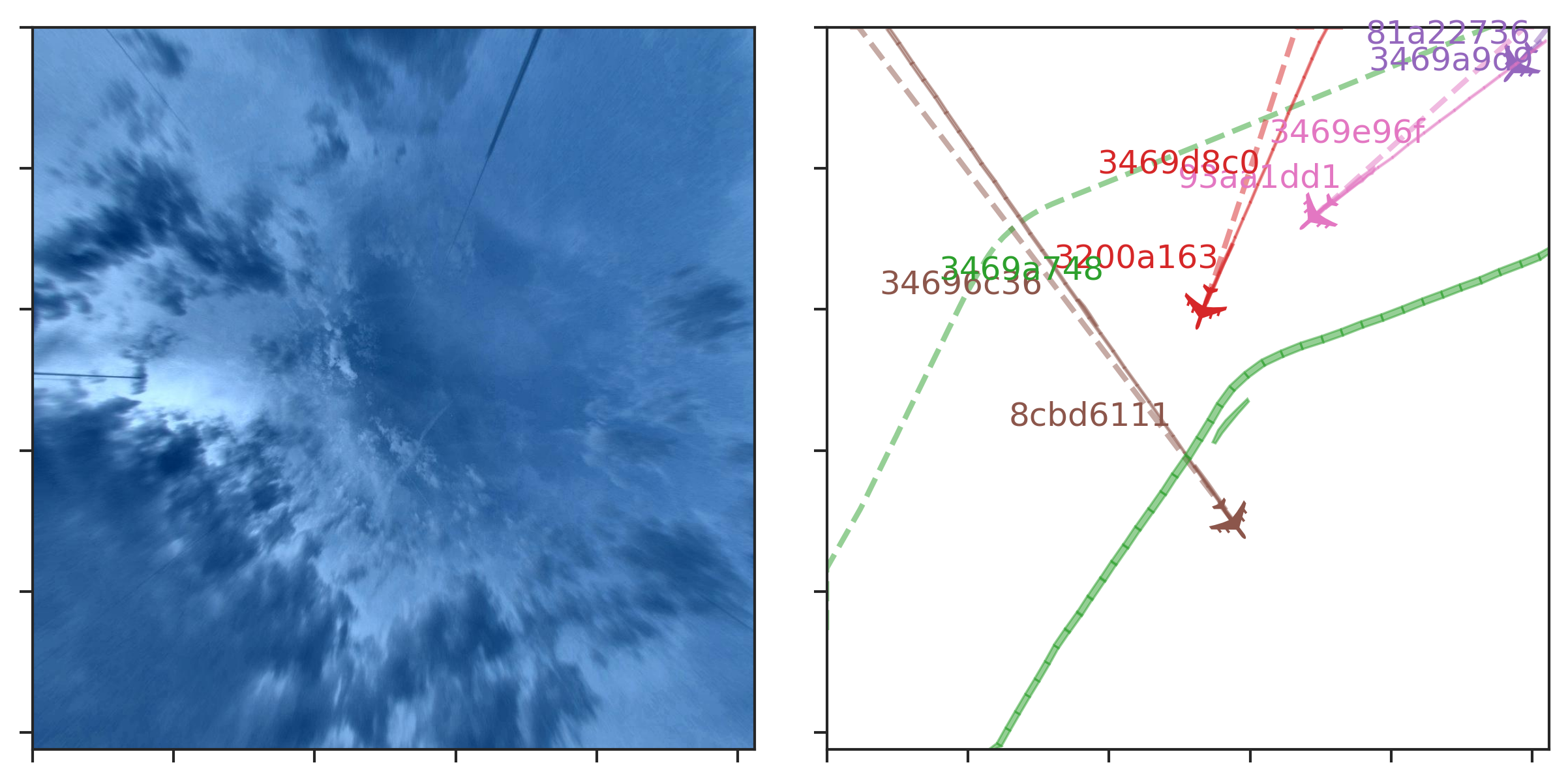}
      \caption{Wrong Attribution}
      \label{fig:attr_wrong}
    \end{subfigure}
    \begin{subfigure}{0.5\textwidth}
      \includegraphics[width=\linewidth]{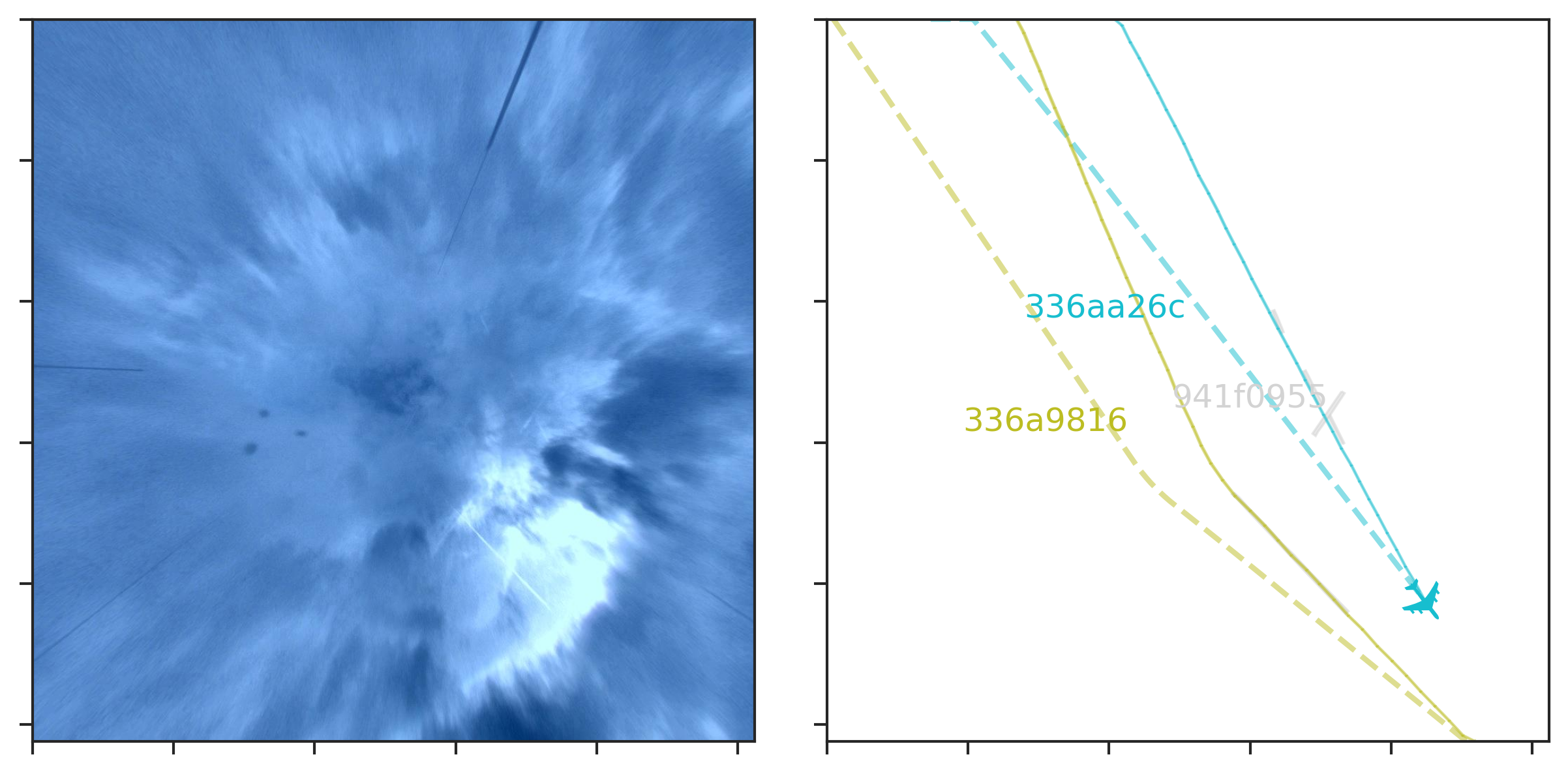}
      \caption{False Attribution}
      \label{fig:attr_false}
    \end{subfigure}
    \caption{Illustrative examples of attributed contrails.}
    \label{fig:attributed}
\end{figure}

Figure~\ref{fig:attributed}a illustrates a textbook case of correct attribution. Each observed contrail is accurately matched to its generating flight: contrail \texttt{b9936bed} to flight \texttt{3453a381}, contrail \texttt{3452e65e} to flight \texttt{b114466}, and contrail \texttt{34543e56} to flight \texttt{0354c9b5}. This example vividly demonstrates the importance of modeling wind advection: strong northwesterly winds shift contrails toward the bottom-right of the frame almost immediately after formation. Without accounting for this drift, aligning contrails to aircraft paths would be challenging.

Figure~\ref{fig:attributed}b shows a wrong attribution, highlighting the limitations of the algorithm under ambiguous conditions. Contrail \texttt{81a22736} was incorrectly assigned to flight \texttt{93aa1dd1}, because an advected contrail from a flight that passed through the same location seconds earlier coincided closely with the observation. This issue could be mitigated by tightening the temporal filtering on contrail formation, for instance. 

Figure~\ref{fig:attributed}c shows a false attribution. In this example, an older contrail (\texttt{941f0955}) is incorrectly assigned to flight \texttt{336aa26c} because the advected path closely aligns with the observed contrail. Such cases illustrate the difficulty of distinguishing persistent contrails from newly formed ones and may also indicate occasional errors in the ground-truth labels.

\begin{figure}[htbp!]
    \centering
    \begin{subfigure}{0.5\textwidth}
      \includegraphics[width=\linewidth]{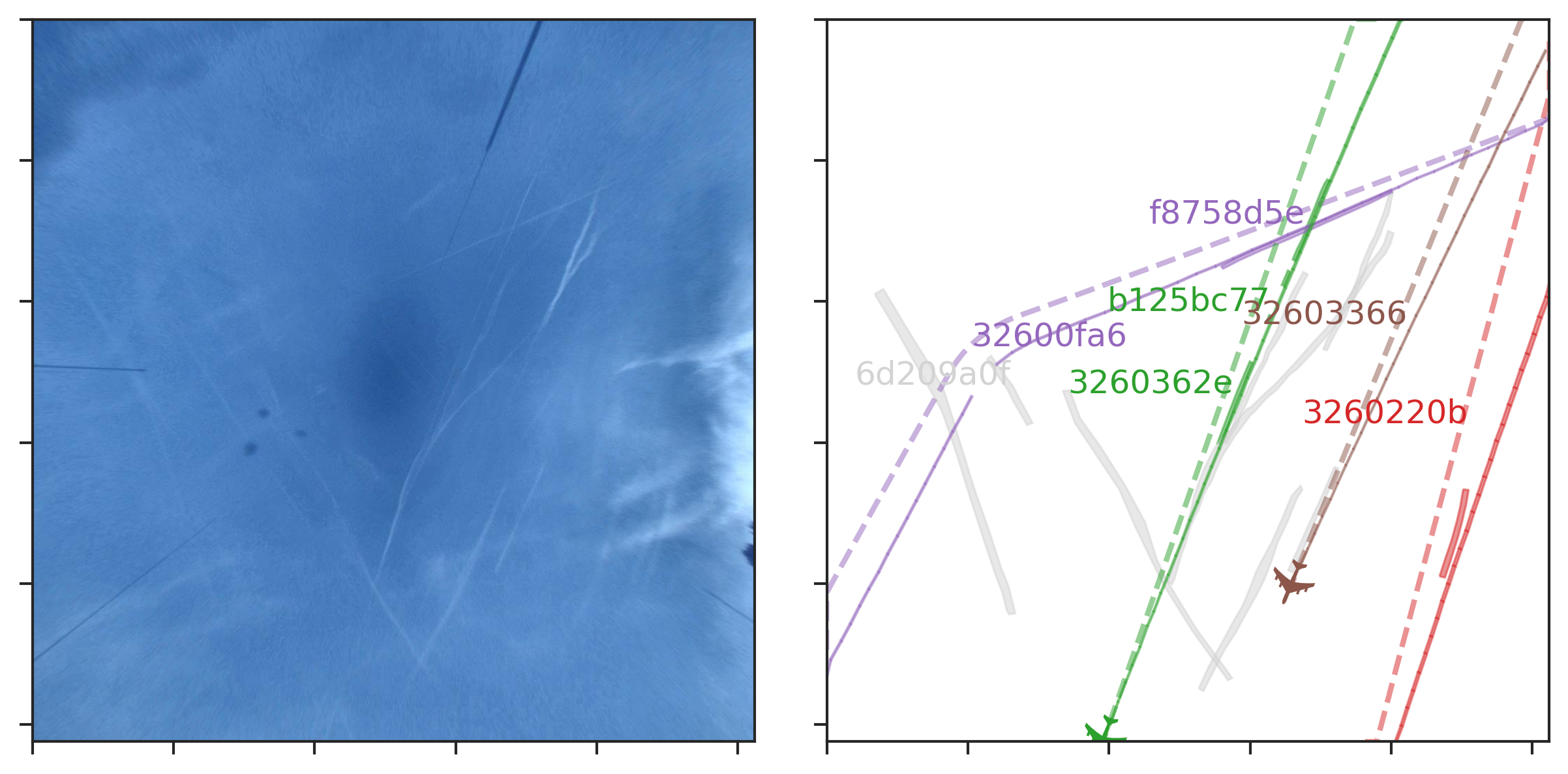}
      \caption{Correct Omission}
      \label{fig:omit_correct}
    \end{subfigure}
    \begin{subfigure}{0.5\textwidth}
      \includegraphics[width=\linewidth]{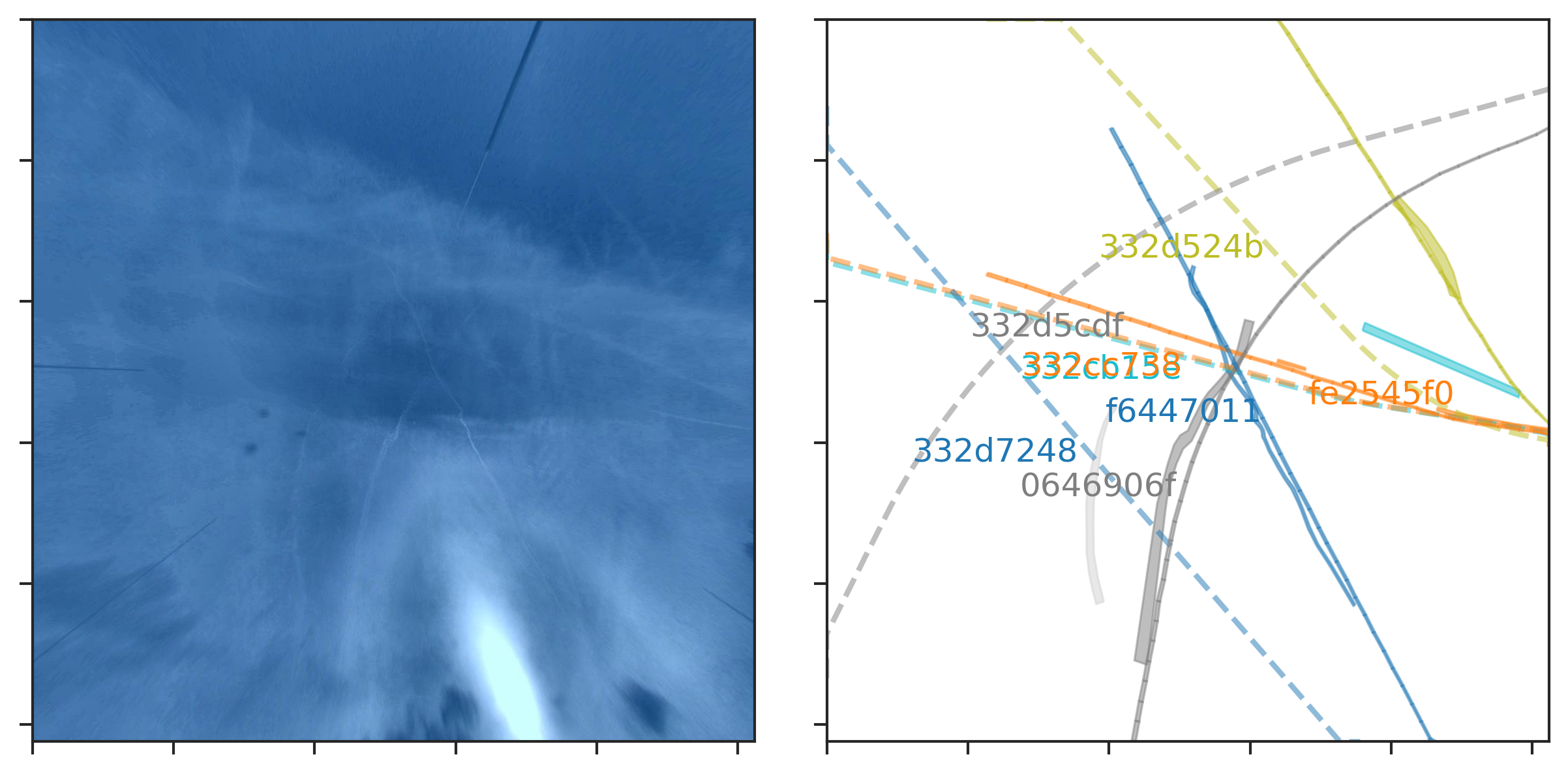}
      \caption{Missed Attribution}
      \label{fig:omit_missed}
    \end{subfigure}
    \caption{Illustrative examples of unattributed contrails.}
    \label{fig:unattributed}
\end{figure}

Figure~\ref{fig:unattributed}a demonstrates a correct omission. Contrail \texttt{6d209a0f} drifts into the camera’s field of view from outside and does not correspond to any advected theoretical contrails in the frame. The algorithm appropriately leaves it unattributed, avoiding false associations.

Finally, Figure~\ref{fig:unattributed}b shows a missed attribution under strong easterly winds. A newly formed contrail (\texttt{0646906f}) was not matched to any flight because the temporal and spatial thresholds prevented a valid candidate from being considered. This example underscores the delicate balance in threshold selection.

\subsection{Aggregated Performance}\label{sec:perf} 

\begin{figure}[htbp!]
    \centering
    \begin{subfigure}{0.5\textwidth}
      \includegraphics[width=\linewidth]{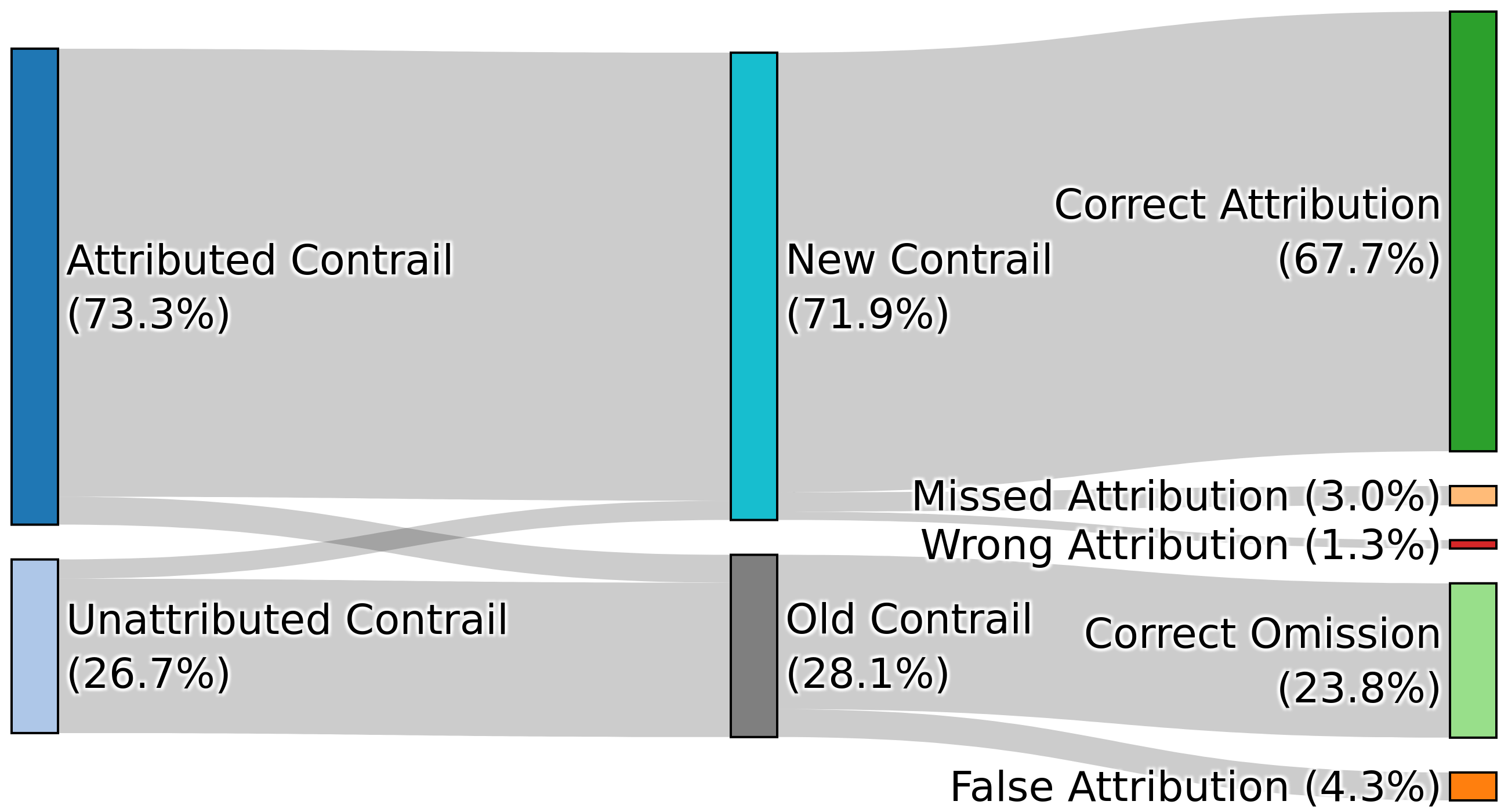}
      \caption{First Attribution}
      \label{fig:sankey_first}
    \end{subfigure}
    \begin{subfigure}{0.5\textwidth}
      \includegraphics[width=\linewidth]{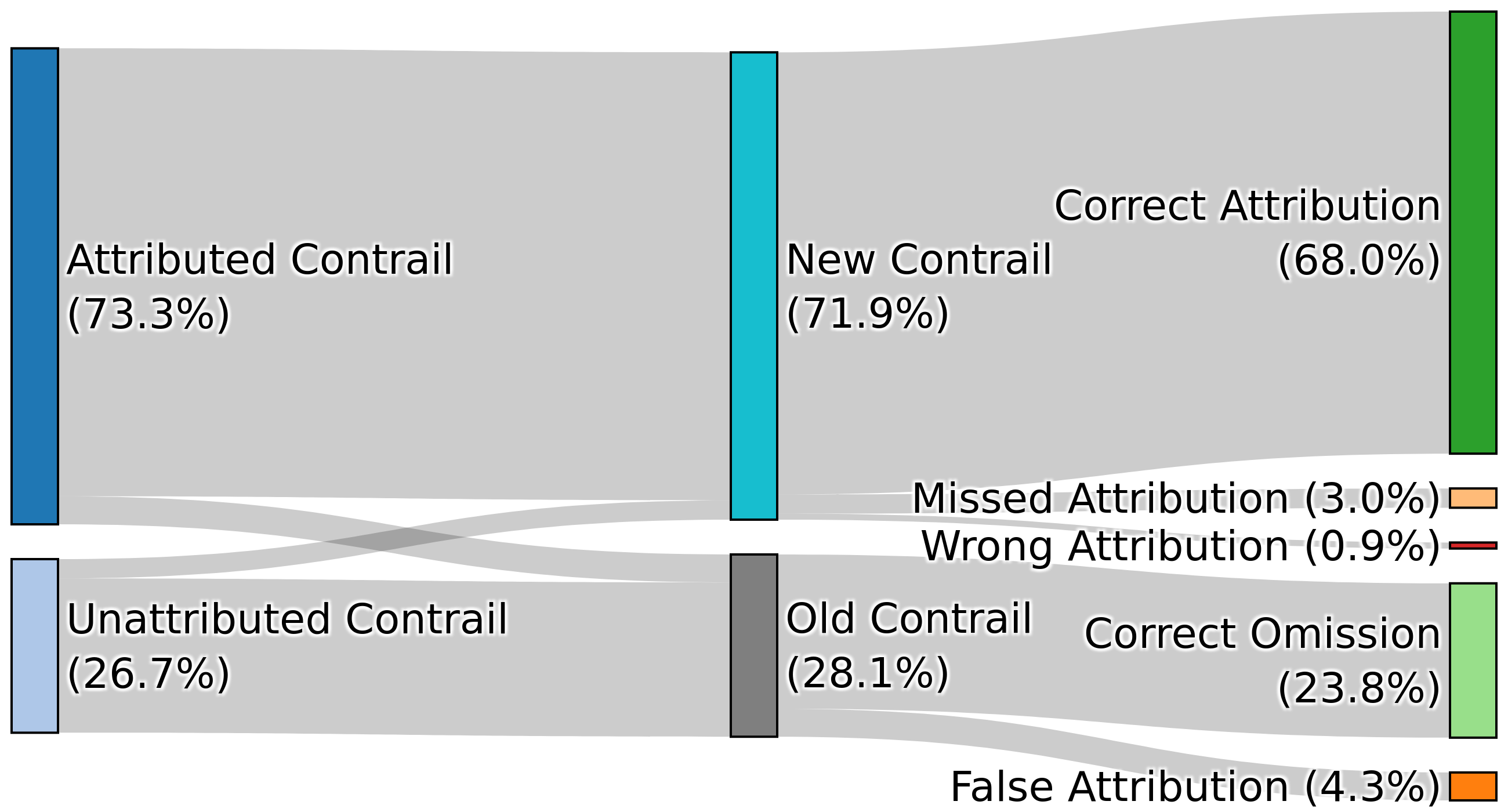}
      \caption{Last Attribution}
      \label{fig:sankey_last}
    \end{subfigure}
    \caption{Sankey diagram illustrating the possible algorithm outcomes.}
    \label{fig:sankey}
\end{figure}

The Sankey diagrams in Figs.~\ref{fig:sankey_first} and~\ref{fig:sankey_last} illustrate the performance of the contrail-to-flight attribution algorithm at two distinct points in time: the first frame in which a contrail becomes visible in the camera imagery, and the final frame in which that contrail remains visible. These diagrams provide an intuitive visualization of how contrails are distributed across attribution outcomes, with the width of each flow proportional to the number of contrails that follow that outcome. 

At the moment of first appearance (Fig.~\ref{fig:sankey_first}), the dataset comprises 71.9\% new contrails and 28.1\% old contrails. The attribution algorithm assigned flights to 73.3\% of all contrails, leaving 26.7\% unattributed.  

Focusing on new contrails, where attribution is crucial, the algorithm correctly identified the generating flight for 67.7\% of the dataset. This translates to a striking 94.2\% correct attribution rate relative to the new-contrail group. Missed attributions, where new contrails remained unattributed, accounted for only 3.0\% of the dataset (4.2\% of new contrails), while wrong attributions (i.e., new contrails assigned to the wrong flight) represented just 1.3\% of the dataset (1.8\% of new contrails). In short, nearly all new contrails were correctly attributed, with only small fractions falling into missed or wrong attribution categories.  

The story is different for old contrails. Most old contrails (23.8\% of the dataset, or 84.7\% of the old-contrail group) were correctly left unattributed. However, 4.3\% of the dataset (15.3\% of old contrails) were falsely assigned to a flight. These false attributions may reflect the algorithm’s occasional difficulty in distinguishing old contrails from newly forming ones. Although this error affects a relatively small portion of the overall dataset, it is proportionally significant within the old-contrail group and constitutes the main challenge for attribution. Some false attributions may also arise from labelling errors, where contrails labelled as old are in fact new but were not linked to their generating flights by human annotators. Given the inherent complexity of this task, no dataset is perfectly labelled. Carefully reviewing these cases and potentially leveraging the algorithm to identify and correct mislabels could be highly valuable. As a result, the algorithm’s true performance for old contrails may be higher than reported.

When performance is examined at the last frame of each contrail’s visibility (Fig.~\ref{fig:sankey_last}), the overall distribution remains largely similar, though some refinements can be observed. Correct attributions rise slightly to 68.0\% of the dataset, corresponding to 94.6\% of new contrails. This increase is modest in absolute terms but demonstrates that the algorithm becomes marginally more reliable when it has access to the full temporal extent of a contrail. Missed attributions remain constant at 3.0\% of the dataset, or 4.2\% of new contrails, indicating that additional temporal evidence does not rescue contrails that were initially left unattributed. Wrong attributions, however, decrease from 1.3\% to 0.9\% of the dataset, corresponding to a reduction from 1.8\% to 1.2\% of new contrails. This decline suggests that temporal accumulation helps disambiguate flight candidates for contrails that might have been incorrectly attributed at the moment of first appearance.

The outcomes for old contrails remain unchanged between first and last attribution. This stability indicates that the algorithm’s errors with old contrails are not mitigated by longer observation windows and temporal evidence. Once an old contrail is mistakenly treated as new and attributed to a flight, additional frames do not alter that decision. Conversely, if an old contrail is correctly omitted at first appearance, it tends to remain so. These results imply that the challenge of distinguishing between old and new contrails is not one of insufficient temporal context but rather of feature design or classification strategy.


Taken together, the two diagrams provide a coherent overview of the algorithm’s performance throughout the contrail lifecycle. In both cases, the algorithm achieves very high attribution accuracy for new contrails. The modest increase in correct attributions and reduction in wrong attributions at the last observation indicate that temporal integration helps disambiguate challenging cases, albeit only slightly. The main limitation of the system is its tendency to falsely attribute a non-negligible fraction of old contrails. Part of this may be due to the inherent complexity of the labelling task: some contrails labelled as old may in fact be new but were not linked to a flight due to occlusion or other visual limitations. The error rate remains roughly constant over time, suggesting that this is a systematic limitation rather than one that can be alleviated by additional temporal evidence.

\section{Conclusions}\label{sec:conclusions}

This work introduced a modular framework for attributing observed contrails in ground-based imagery to their source flight, based on geometric and temporal reasoning. 

From a dataset-level perspective, correct attributions account for roughly two-thirds of all contrails, while correct omissions make up about a quarter. The remaining errors, comprising missed attributions, wrong attributions, and false attributions, collectively represent approximately 9\% of the dataset. However, examining performance within the relevant groups highlights the algorithm’s strengths. For new contrails, attribution precision reaches around 92–93\%, with a recall of approximately 94–95\%. For old contrails, correct omissions are frequent but not perfect: about one in six old contrails is falsely attributed to a flight. Some of these errors may reflect genuine algorithm limitations, but a fraction may also stem from labelling errors, where contrails labelled as old are in fact new but were not linked to a generating flight due to occlusion, for instance.

The framework we propose lays the groundwork for several promising extensions. A natural next step is the incorporation of contrail altitude estimation, either directly from imagery or by combining visual data with meteorological or ADS-B-derived information. Altitude cues would help disambiguate overlapping flight paths and improve spatial matching, especially in densely trafficked airspace. 

A particularly compelling extension lies in combining ground-based and satellite imagery to track contrails beyond the camera's field of view. Ground cameras offer high-resolution detections and accurate attribution at the early stages of contrail formation. By detecting the same contrail in both ground and satellite imagery, and using the ground-based attribution as a reference, one can establish contrail-to-contrail correspondences across modalities. This enables continued tracking of the contrail's evolution even after it exits the ground camera’s view, capturing the full spatial and temporal extent of the contrail. Such an approach could unlock powerful new capabilities in monitoring persistent contrails and assessing their climate impact over time, particularly as they spread into cirrus-like formations far from their origin.

\bibliographystyle{IEEEtran}
\bibliography{reference}

\begin{thebibliography}{10}
\providecommand{\url}[1]{#1}
\csname url@samestyle\endcsname
\providecommand{\newblock}{\relax}
\providecommand{\bibinfo}[2]{#2}
\providecommand{\BIBentrySTDinterwordspacing}{\spaceskip=0pt\relax}
\providecommand{\BIBentryALTinterwordstretchfactor}{4}
\providecommand{\BIBentryALTinterwordspacing}{\spaceskip=\fontdimen2\font plus
\BIBentryALTinterwordstretchfactor\fontdimen3\font minus \fontdimen4\font\relax}
\providecommand{\BIBforeignlanguage}[2]{{%
\expandafter\ifx\csname l@#1\endcsname\relax
\typeout{** WARNING: IEEEtran.bst: No hyphenation pattern has been}%
\typeout{** loaded for the language `#1'. Using the pattern for}%
\typeout{** the default language instead.}%
\else
\language=\csname l@#1\endcsname
\fi
#2}}
\providecommand{\BIBdecl}{\relax}
\BIBdecl

\bibitem{lee2021contribution}
D.~S. Lee, D.~W. Fahey, A.~Skowron, M.~R. Allen, U.~Burkhardt, Q.~Chen, S.~J. Doherty, S.~Freeman, P.~M. Forster, J.~Fuglestvedt \emph{et~al.}, ``The contribution of global aviation to anthropogenic climate forcing for 2000 to 2018,'' \emph{Atmospheric Environment}, vol. 244, p. 117834, 2021.

\bibitem{teoh2023global}
R.~Teoh, Z.~Engberg, U.~Schumann, C.~Voigt, M.~Shapiro, S.~Rohs, and M.~Stettler, ``Global aviation contrail climate effects from 2019 to 2021,'' \emph{EGUsphere}, vol. 2023, pp. 1--32, 2023.

\bibitem{borella2024importance}
A.~Borella, O.~Boucher, K.~P. Shine, M.~Stettler, K.~Tanaka, R.~Teoh, and N.~Bellouin, ``The importance of an informed choice of co 2-equivalence metrics for contrail avoidance,'' \emph{Atmospheric Chemistry and Physics}, vol.~24, no.~16, pp. 9401--9417, 2024.

\bibitem{schumann2012cocip}
\BIBentryALTinterwordspacing
U.~Schumann, ``A contrail cirrus prediction model,'' \emph{Geoscientific Model Development}, vol.~5, no.~3, pp. 543--580, 2012. [Online]. Available: \url{https://gmd.copernicus.org/articles/5/543/2012/}
\BIBentrySTDinterwordspacing

\bibitem{fritz2020role}
T.~M. Fritz, S.~D. Eastham, R.~L. Speth, and S.~R. Barrett, ``The role of plume-scale processes in long-term impacts of aircraft emissions,'' \emph{Atmospheric Chemistry and Physics}, vol.~20, no.~9, pp. 5697--5727, 2020.

\bibitem{gierens2020uncertainty}
K.~Gierens, S.~Matthes, and S.~Rohs, ``How well can persistent contrails be predicted?'' \emph{Aerospace}, vol.~7, no.~12, 2020.

\bibitem{chevallier2023linear}
R.~Chevallier, M.~Shapiro, Z.~Engberg, M.~Soler, and D.~Delahaye, ``Linear contrails detection, tracking and matching with aircraft using geostationary satellite and air traffic data,'' \emph{Aerospace}, vol.~10, no.~7, p. 578, 2023.

\bibitem{van2025contrail}
J.~Van~Huffel, R.~Ehrmanntraut, and D.~Croes, ``Contrail detection and classification using computer vision with ground-based cameras,'' in \emph{2025 Integrated Communications, Navigation and Surveillance Conference (ICNS)}.\hskip 1em plus 0.5em minus 0.4em\relax IEEE, 2025, pp. 1--6.

\bibitem{jarry_2025_15743988}
\BIBentryALTinterwordspacing
G.~Jarry, P.~Very, F.~Ballerini, and R.~Dalmau, ``Gvccs : Ground visible camera contrail sequences,'' Jul. 2025. [Online]. Available: \url{https://doi.org/10.5281/zenodo.15743988}
\BIBentrySTDinterwordspacing

\bibitem{jarry2025gvccs}
G.~Jarry, R.~Dalmau, P.~Very, F.~Ballerini, and S.-D. Bocu, ``Gvccs: A dataset for contrail identification and tracking on visible whole sky camera sequences,'' \emph{Earth System Science Data Discussions}, vol. 2025, pp. 1--30, 2025.

\bibitem{itcovitz2024attribution}
J.~{Itcovitz}, A.~{Sarna}, M.~{Stettler}, and V.~{Meijer}, ``{Flight-Matched Contrail Observations from GOES - Insights from Aircraft-Engine Analysis},'' in \emph{AGU Fall Meeting Abstracts}, vol. 2024, Dec. 2024, pp. A51M--1867.

\bibitem{geraedts2024contrails}
S.~Geraedts, E.~Brand, T.~Dean, S.~Eastham, C.~Elkin, Z.~Engberg, U.~Hager, I.~Langmore, K.~McCloskey, J.~Ng, J.~Platt, T.~Sankar, A.~Sarna, M.~Shapiro, and N.~Goyal, ``A scalable system to measure contrail formation on a per-ﬂight basis,'' \emph{Environmental Research Communications}, vol.~6, 01 2024.

\bibitem{sarna2025benchmarking}
A.~Sarna, V.~Meijer, R.~Chevallier, A.~Duncan, K.~McConnaughay, S.~Geraedts, and K.~McCloskey, ``Benchmarking and improving algorithms for attributing satellite-observed contrails to flights,'' \emph{EGUsphere}, vol. 2025, pp. 1--58, 2025.

\bibitem{low2025ground}
J.~Low, R.~Teoh, J.~Ponsonby, E.~Gryspeerdt, M.~Shapiro, and M.~E. Stettler, ``Ground-based contrail observations: comparisons with reanalysis weather data and contrail model simulations,'' \emph{Atmospheric Measurement Techniques}, vol.~18, no.~1, pp. 37--56, 2025.

\bibitem{riggi2023ai}
E.~Riggi-Carrolo, T.~Dubot, C.~Sarrat, and J.~Bedouet, ``Ai-driven identification of contrail sources: Integrating satellite observation and air traffic data,'' \emph{Journal of Open Aviation Science}, vol.~1, no.~2, 2023.

\bibitem{duda2024clusters}
\BIBentryALTinterwordspacing
D.~P. Duda, P.~Minnis, L.~Nguyen, and R.~Palikonda, ``A case study of the development of contrail clusters over the great lakes,'' \emph{Journal of the Atmospheric Sciences}, vol.~61, no.~10, pp. 1132 -- 1146, 2004. [Online]. Available: \url{https://journals.ametsoc.org/view/journals/atsc/61/10/1520-0469_2004_061_1132_acsotd_2.0.co_2.xml}
\BIBentrySTDinterwordspacing

\end{thebibliography}

\end{document}